%% file: ccl2022-en.tex
\pdfoutput=1

\documentclass[11pt]{article}
\usepackage[hyperref]{ccl2022-en}
\usepackage{times}
\usepackage{url}
\usepackage{latexsym}
\usepackage{fancyhdr}
\usepackage{multicol}
\usepackage{multirow}
\usepackage{graphicx}
\usepackage{subcaption}
\usepackage{placeins}
\usepackage{makecell}
\usepackage{amsmath}
\usepackage{float}
\usepackage{lettrine}
\usepackage{tikz}
\usetikzlibrary{pgfplots.groupplots}
\usepackage{pgfplots}\pgfplotsset{compat=1.17} 

\title{To Adapt or to Fine-tune: A Case Study on Abstractive Summarization}

\author{Zheng Zhao \qquad\qquad
  Pinzhen Chen \\
  School of Informatics, University of Edinburgh\\
  \texttt{\{zheng.zhao,pinzhen.chen\}@ed.ac.uk}}

\date{}

\begin{document}
\maketitle
\begin{abstract}
Recent advances in the field of abstractive summarization leverage pre-trained language models rather than train a model from scratch. However, such models are sluggish to train and accompanied by a massive overhead. Researchers have proposed a few lightweight alternatives such as smaller adapters to mitigate the drawbacks. Nonetheless, it remains uncertain whether using adapters benefits the task of summarization, in terms of improved efficiency without an unpleasant sacrifice in performance. In this work, we carry out multifaceted investigations on fine-tuning and adapters for summarization tasks with varying complexity: language, domain, and task transfer. In our experiments, fine-tuning a pre-trained language model generally attains a better performance than using adapters; the performance gap positively correlates with the amount of training data used. Notably, adapters exceed fine-tuning under extremely low-resource conditions. We further provide insights on multilinguality, model convergence, and robustness, hoping to shed light on the pragmatic choice of fine-tuning or adapters in abstractive summarization. 
\end{abstract}

\section{Introduction}
\label{intro}

In the current era of research, using large pre-trained language models (PLM) and fine-tuning these models on a downstream task yields dominating results in many tasks \cite{devlin-etal-2019-bert,lewis-etal-2020-bart,raffel-etal-2020-exploring,brown-etal-2020-language}. The scope of our work is on abstractive summarization, which is the task of generating a concise and relevant summary given a long document. Recent works have demonstrated the success of fine-tuning PLMs on summarization \cite{liu-lapata-2019-text,zhang-etal-2020-pegasus,rothe-etal-2020-leveraging}. Nonetheless, such a paradigm becomes increasingly expensive with the ever-growing sizes of PLMs, since both the training time and space requirement increase along with the number of parameters. The issue becomes more severe when multiple languages or domains are introduced, as separate models need to be trained and saved depending on the setup. 

\newcite{houlsby-etal-2019-parameter} proposed lightweight adapters as an alleviation of the large overhead of fine-tuning PLM on a downstream task. While many researchers have followed and adopted their idea, experiments are rarely done on summarization; from both quantitative and qualitative perspectives, it remains a myth of which direction one should pick in practice. In this work, we perform a thorough exploration of using adapters with a PLM on the task of abstractive summarization by examining different scenarios.

Our experiments are designed along three dimensions: 1) languages involved: monolingual, cross-lingual, and multilingual; 2) data availability: high, medium, low, and scarce; 3) knowledge being transferred: languages, domains as well as tasks. Through comprehensive experimental results, we demonstrate that with a realistic availability of resources, fine-tuning a PLM is superior to using adapters for the purpose of obtaining the best text quality. However, the game changes under low-resource settings: adapters have shown better, if not, on par performances compared to fine-tuning, especially in domain adaption.

\section{Related Work}\label{sec:related-work}

Fine-tuning a PLM with downstream task-specific objectives is a useful paradigm. It not only speeds up training, but also transfers the knowledge from abundant pre-training data to lower-resourced tasks. Whilst it has been proven successful in the field of summarization \cite{ladhak-etal-2020-wikilingua,zhang-etal-2020-pegasus,zou-etal-2020-pre,rothe-etal-2020-leveraging}, this strategy requires optimizing and updating all parameters in the fine-tuned model, and is particularly expensive when a number of (sub-)tasks need to be approached. 

To mitigate these problems, \newcite{houlsby-etal-2019-parameter} proposed to insert small neural modules named ``adapters'' to each layer of the PLM sequentially, and only update the adapters during fine-tuning while freezing most of the PLM parameters. When dealing with different sub-tasks -- languages, domains, etc. -- it is especially storage-efficient as only adapter weights need to be saved instead of the whole fine-tuned model. Several adapter architectures have been designed since then. \newcite{pfeiffer-etal-2020-mad} suggested simply placing adapters after the feed-forward block in each layer of the PLM, instead of adding adapters after both the multi-head attention and feed-forward block as proposed in the original work. Apart from adding adapters sequentially, \newcite{he-etal-2022-towards} designed an adapter that is parallel to the PLM. 

Recent research that had utilized adapters in the task of summarization, argued that the low availability of opinion summarization datasets often leads to the standard fine-tuning method overfitting on tiny datasets \cite{brazinskas-2022-adasum}. Thus, they presented an efficient few-shot fine-tuning method based on adapters for opinion summarization. They added adapters to pre-trained models, trained the adapters on a large unlabelled customer reviews dataset, then fine-tuned them on the human-annotated corpus. Their method outperformed standard fine-tuning methods on various datasets. In addition, they showed that the proposed method can generate better-organized summaries with improved coherence and fewer redundancies in the case of summary personalization. \newcite{chen-shuai-2021-meta} created a meta-transfer learning framework for low-resource abstractive summarization, aiming to leverage pre-trained knowledge to improve the performance of the target corpus with limited examples. They inserted adapter modules into their model to perform meta-learning and leverage pre-trained knowledge simultaneously. Their methods are particularly effective under manually constructed low-resource settings on various summarization datasets with diverse writing styles and forms.

In comparison, our work investigates fine-tuning and using adapters in summarization, by comparing the performance of models using the fine-tuning strategy with models using adapters in the case of language adaptability, data availability, and knowledge transfer. For language adaptability, we examine the case of monolingual, cross-lingual, and multilingual summarization. For data availability, we study models trained under low, medium, and high resource scenarios. Lastly, for knowledge transfer, we investigate several factors: languages, domains, and tasks. To the best of our knowledge, adapters have not been tested in these scenarios.

\section{Methodology}

\subsection{Method overview}
Our aim is to study two fine-tuning variants for summarization under several settings using a PLM: the \textit{fine-tuning} paradigm, and the \textit{adapter} strategy. Fine-tuning initializes a PLM from a pre-trained checkpoint, then trains and updates the whole model on a summarization dataset. On the other hand, the adapter strategy also initializes a PLM from a pre-trained checkpoint, with adapter modules then inserted into the model. During training, we only update the adapter, the layer normalization parameters, and the final output layer.  

We use mBART \cite{liu-etal-2020-multilingual-denoising} as our backbone PLM for settings involving non-English languages. It is a sequence-to-sequence model pre-trained on large-scale monolingual corpora in 25 languages, with a denoising autoencoding objective. The model is designed to do multilingual machine translation tasks. After training it on a summarization dataset, the model is capable of doing monolingual, cross-lingual, and multilingual summarization. For English-only settings, we use BART \cite{lewis-etal-2020-bart} as the PLM. Similar to mBART, BART is also a sequence-to-sequence model pre-trained on large-scale corpora with denoising autoencoder architecture. 

\begin{figure*}[t]
    \centering
    \includegraphics[width=\linewidth]{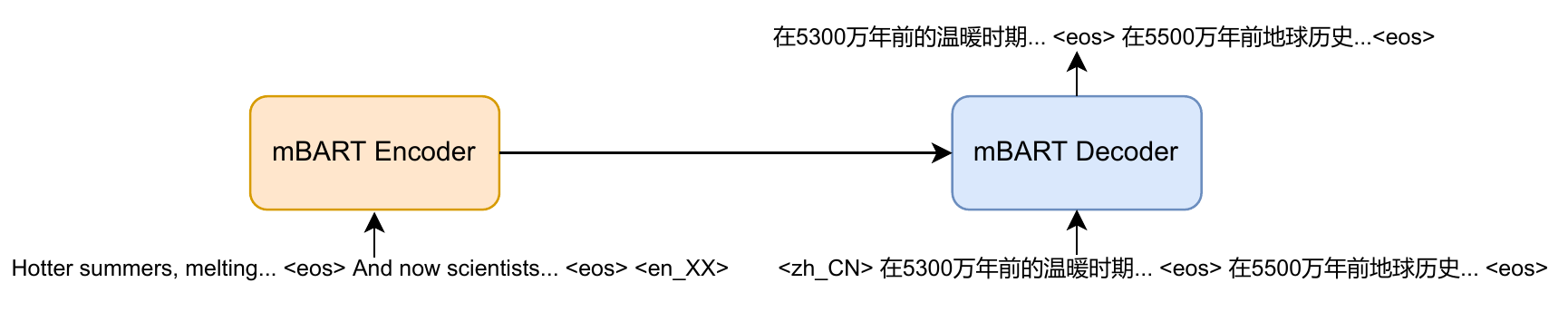}
    \caption{An illustration of our mBART based model for cross-lingual summarization from English to Chinese.}
    \label{fig:model_architecture}
\end{figure*}

We have two kinds of models: mBART-FT which employs the fine-tune strategy, and mBART-Adapt which uses the adapt strategy. In order to recognise the source and target languages, following \newcite{liu-etal-2020-multilingual-denoising}, our models take a special separator token between each sentence, a language code token at the end of the source document, and at the beginning of the target summary. We provide a cross-lingual demonstration for our model in Figure~\ref{fig:model_architecture}. In addition, we propose BART-FT and BART-Adapt which use the fine-tune strategy and the adapt strategy, respectively.

\subsection{Adapter variants}
\begin{figure*}[ht]
     \centering
     \begin{subfigure}[b]{0.4\textwidth}
         \centering
         \includegraphics[width=\textwidth]{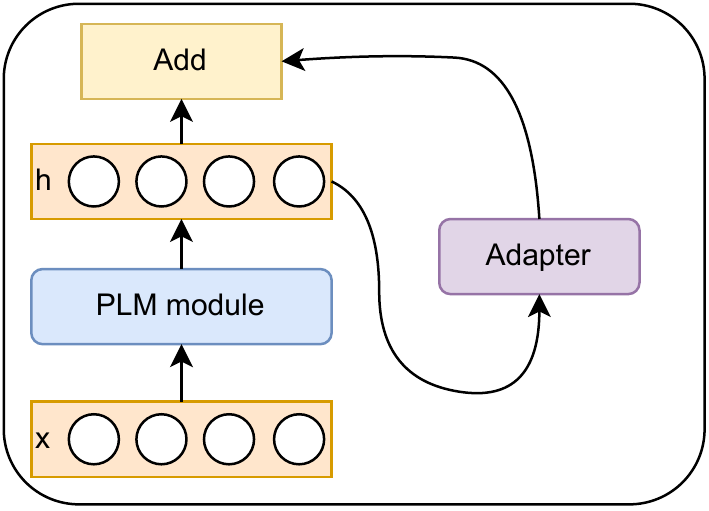}
         \caption{A sequential adapter}
         \label{fig:adapter_seq}
     \end{subfigure}
     \hspace{10ex}
     \begin{subfigure}[b]{0.4\textwidth}
         \centering
         \includegraphics[width=\textwidth]{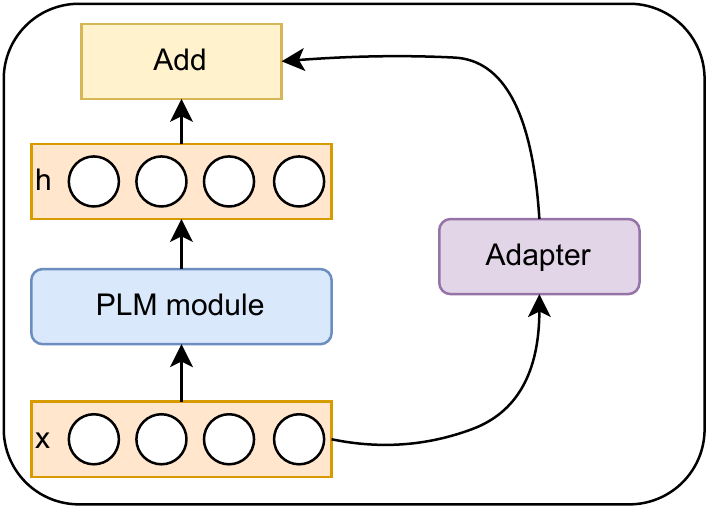}
         \caption{A parallel adapter}
         \label{fig:adapter_parallel}
     \end{subfigure}
        \caption{An illustration of adapter variants, adapted from \newcite{he-etal-2022-towards}. ``PLM module'' represents a certain sub-layer of the PLM (e.g. attention or feed-forward layer) that is frozen.}
        \label{fig:adapter_variants}
\end{figure*}

As mentioned earlier, there are various adapter variants. We experiment with two variants: one with sequential connections \cite{houlsby-etal-2019-parameter}, and one with parallel connections \cite{he-etal-2022-towards}. We display an illustration of these variants in Figure~\ref{fig:adapter_variants}. After trying out different learning rates and reduction factors (the ratio between PLM's hidden dimension and adapter's bottleneck dimension), we discover that sequential adapters always outperform the parallel ones in our tasks. Thus we use \newcite{houlsby-etal-2019-parameter}'s sequential adapter for all of our mBART/BART-Adapt models.

\subsection{Evaluation}
The evaluation metrics are F1 scores of ROUGE-1/2/L \cite{lin-2004-rouge}. Since we deal with multiple languages, we use the multilingual ROUGE implemented in a previous paper.\footnote{\url{https://github.com/csebuetnlp/xl-sum/tree/master/multilingual_rouge_scoring}} We stick to the toolkit's default settings, e.g., sentence segmentation and word stemming.

\begin{table}[bht]
\centering
\begin{tabular}{|c|c|r|c|}
\hline
   \textbf{Dataset} & \textbf{Language} & \textbf{train\ /\ valid\ /\ test} & \textbf{Source} \\
\hline
 \multirow{2}{*}{NCLS}  & zh$\rightarrow$en  & 1.7m\ /\ 3.0k\ /\ 3.0k & Sina Weibo \\
                        & en$\rightarrow$zh  & 365k\ /\ 3.0k\ /\ 3.0k & CNN/Daily Mail \\
                             
\hline
 \multirow{3}{*}{\makecell{Wiki-\\Lingua}} & en$\rightarrow$ar & 20.4k\ /\ 2.9k\ /\ 5.8k & \multirow{3}{*}{wikiHow} \\
 & en$\rightarrow$vi & 13.7k\ /\ 2.0k\ /\ 3.9k & \\
 & en$\leftrightarrow$ja & 8.9k\ /\ 1.3k\ /\ 2.5k & \\
%  \cline{2-3}
%  & zh$\rightarrow$en & k\ /\ k\ /\ k & \\
\hline
 \multirow{5}{*}{XL-Sum} & gu & 9.1k\ /\ 1.1k\ /\ 1.1k & \multirow{5}{*}{BBC}  \\
 & fr & 8.7k\ /\ 1.1k\ /\ 1.1k & \\
 & ne & 5.8k\ /\ 0.7k\ /\ 0.7k & \\
 & ko & 4.4k\ /\ 0.6k\ /\ 0.6k & \\
 & si & 3.2k\ /\ 0.5k\ /\ 0.5k & \\
\hline
\end{tabular}
\caption{Statistics of datasets and languages for the language adaption experiment.}
\label{tab:data}
\end{table}

\section{Language Experiments}
\label{sec:4-exp}
\subsection{Experimental setup} 
We test our proposed paradigm on \text{NCLS}\footnote{\url{https://github.com/znlp/ncls-corpora}}, \text{WikiLingua}\footnote{\url{https://github.com/esdurmus/wikilingua}}, and \text{XL-Sum}\footnote{\url{https://github.com/csebuetnlp/xl-sum}} datasets particularly designed for cross-lingual and multilingual summarization \cite{zhu-etal-2019-ncls,ladhak-etal-2020-wikilingua,hasan-etal-2021-xl}. These datasets are either machine-translated or crawled from the web.

NLCS is built by machine-translating an existing English (en) dataset (CNN/Daily Mail, by \newcite{nallapati-etal-2016-abstractive}) to Chinese (zh), and vice versa (Sina Weibo, by \newcite{hu-etal-2015-lcsts}). A translated document is only kept if its round-trip translation reaches a certain threshold score. Plain translations and human-corrected translations are supplied as separate test sets; we use the human-corrected set in this work. WikiLingua is constructed by extracting and aligning article-summary pairs from wikiHow. We experiment with three languages that resemble medium and low-resource scenarios: Arabic (ar), Vietnamese (vi), and Japanese (ja). 

Different from the cross-lingual datasets, XL-Sum is monolingual. It consists of professionally annotated article-summary pairs from BBC in many languages. The datasets come in various sizes for a number of languages, as shown in Table~\ref{tab:data}. This dataset allows for multilingual experiments since the data come from the same domain and are not centred on English. We experiment on five low-resource languages: Gujarati (gu), French (fr), Nepali (ne), Korean (ko), and Sinhala (si). For the monolingual scenario, we directly use the monolingual summarization data to train the model. For the cross-lingual setting, since machine translation is a cross-lingual task, we also directly train the model using the cross-lingual summarization data. Lastly, in a multilingual configuration, we simply mix summarization data in different languages, and train the model using the mixed data.

\begin{table}[ht]
% \hfill
\begin{subtable}[ht] {\textwidth}
\centering
\begin{tabular}{|c|c|c|c|c|c|c|}
\hline
\multirow{2}{*}{Lang.}
 & \multicolumn{3}{c|}{mBART-FT} 
 & \multicolumn{3}{c|}{mBART-Adapt}\\
\cline{2-7}
 & R1 & R2 & RL & R1 & R2 & RL  \\
\hline
 zh$\rightarrow$en & \textbf{46.46} & \textbf{30.18} & \textbf{42.26} & 41.41 & 22.73 & 36.56 \\
 en$\rightarrow$zh & \textbf{45.22} & \textbf{22.49} & \textbf{34.38} & 40.74 & 16.83 & 29.27 \\
\hline
\end{tabular}
\caption{High-resource, NCLS.}
\label{tab:ncls}
\end{subtable}
 %\hfill
 \\

\begin{subtable}[ht] {\textwidth}
\centering
\begin{tabular}{|c|c|c|c|c|c|c|}
\hline
\multirow{2}{*}{Lang.}
 & \multicolumn{3}{c|}{mBART-FT} 
 & \multicolumn{3}{c|}{mBART-Adapt}\\
\cline{2-7} & R1 & R2 & RL & R1 & R2 & RL  \\
\hline
 en$\rightarrow$ar & \textbf{25.85} & \textbf{7.35} & \textbf{21.01} & 24.68 & 7.26 & 20.40 \\
 en$\rightarrow$vi & \textbf{33.63} & \textbf{15.17} & \textbf{26.65} & 30.98 & 13.94 & 24.59 \\
 en$\rightarrow$ja & \textbf{35.70} & \textbf{12.34} & \textbf{28.34} & 34.06 & 11.43 & 27.08 \\
 ja$\rightarrow$en & \textbf{35.24} & \textbf{12.38} & \textbf{28.09} & 33.14 & 11.54 & 26.46 \\
\hline
\end{tabular}
\caption{Medium and low-resource, WikiLingua.}
\label{tab:wikilingua}
 \end{subtable}

 \caption{Results for cross-lingual summarization.}
\label{tab:crosslingual}
\end{table}

\begin{table}[ht]
\centering
\setlength{\tabcolsep}{0.97ex}
\begin{tabular}{|c|c|c|c|c|c|c|c|c|c|c|c|c|}
\hline
\multirow{3}{*}{Lang.} 
 & \multicolumn{6}{c|}{Multilingual} 
 & \multicolumn{6}{c|}{Monolingual} \\
\cline{2-13}
 & \multicolumn{3}{c|}{mBART-FT} 
 & \multicolumn{3}{c|}{mBART-Adapt}
 & \multicolumn{3}{c|}{mBART-FT} 
 & \multicolumn{3}{c|}{mBART-Adapt} \\
\cline{2-13}
 & R1 & R2 & RL & R1 & R2 & RL & R1 & R2 & RL & R1 & R2 & RL \\
\hline
 gu   & \textbf{20.18} & \textbf{6.96} & \textbf{18.09} & 20.12 & 6.82 & 17.99 & \textbf{20.23} & \textbf{6.43} & \textbf{17.67} & 19.20 & 5.95 & 16.96 \\
 fr   & \textbf{33.53} & \textbf{14.37} & \textbf{26.11} & 33.44 & 14.01 & 25.63 & \textbf{33.29} & \textbf{13.68} & \textbf{25.13} & 32.37 & 13.02 & 24.73 \\
 ne   & \textbf{24.70} & \textbf{9.52} & \textbf{22.23} & 23.26 & 8.55 & 20.94 & \textbf{24.06} & \textbf{9.05} & \textbf{21.62} & 23.31 & 8.36 & 21.01 \\
 ko   & 17.73 & \textbf{8.76} & 16.27 & \textbf{18.82} & 8.12 & \textbf{17.23} & \textbf{19.73} & 9.12 & \textbf{18.07} & 19.05 & \textbf{9.24} & 17.73 \\
 si   & \textbf{26.95} & \textbf{13.51} & \textbf{22.36} & 25.68 & 12.69 & 21.80 & \textbf{25.59} & 12.25 & \textbf{21.92} & 24.99 & \textbf{12.30} & 21.44 \\
\hline
\end{tabular}
\caption{Results for low-resource multilingual and monolingual summarization on XL-Sum.}
\label{tab:xl-sum}
\end{table}

Our experiments are based on a public mBART checkpoint\footnote{\url{https://huggingface.co/facebook/mbart-large-cc25}}. We use the adapter from \newcite{houlsby-etal-2019-parameter}. Fine-tuning an mBART model updates around 610M parameters in total; the addition of adapters introduces 50M parameters, yet only this 8\% are being optimized during training. We use the Adam optimizer for training \cite{KingmaB14}, with a learning rate of 1e-5 for mBART, and 1e-4 for mBART with adapters. We set the adapter reduction factor to 2, which means that the bottleneck dimension in an adapter is half of the hidden dimension in mBART. We perform hyperparameter searches on the following: learning rate and reduction factor, and monitor ROUGE scores on the validation set to select the best value. We provide further details of the grid search in Appendix~\ref{sec:appendix-hyper-parameters}. 

All models are trained on 4 NVIDIA A100 GPUs with a batch size of 12 on NCLS, and 4 on WikiLingua and XL-Sum. The model convergence time is from 1 to 30 hours depending on the dataset used. We use PyTorch \cite{NEURIPS2019_9015} for our model implementation. We use the Huggingface library \cite{wolf2019huggingface} and AdapterHub \cite{pfeiffer-etal-2020-adapterhub} for mBART and adapter implementation.

\subsection{Results}

We first provide results on high-recourse cross-lingual summarization on NCLS in Table~\ref{tab:ncls}. We can see that mBART-FT achieves significantly higher ROUGE scores than mBART-Adapt in both Chinese-to-English as well as English-to-Chinese settings. We then list result numbers on medium and low-recourse cross-lingual summarization on WikiLingua in Table~\ref{tab:wikilingua}. Similar to the behaviour under the high-resource setting, mBART-FT consistently achieves better ROUGE performance than mBART-Adapt, regardless of the source or target languages. However, we spot that the difference in ROUGE scores is smaller for language pairs with lower resources, which suggests a positive correlation between the gap in performance and training data availability.

In Table~\ref{tab:xl-sum}, we show results of both multilingual (left) and monolingual (right) summarization on XL-Sum. In a multilingual setup, a single model is trained on five languages. whereas in a monolingual setup, five individual models are trained on the five languages separately. We can first see that mBART-FT generally surpasses mBART-Adapt, in both multilingual and monolingual setups. In addition, multilingual models generally outperform monolingual models by a small margin. This behaviour is corroborated by \newcite{hasan-etal-2021-xl}'s work that mixing multiple languages altogether during training can result in a positive transfer among them \cite{conneau-etal-2020-unsupervised}.

It is straightforward from our work, that, for summarization tasks with high data availability, it is not worth trading performance for efficiency with adapters. For low-resource scenarios, adapters achieve similar results as fine-tuning, and can therefore be a convenient choice for fast training and compact disk storage. When multiple low-resource languages are concerned, especially if they are related languages, it might be beneficial to build a multilingual model instead of individual monolingual models.

\subsection{Convergence}
To measure the convergence difference between mBART-FT and mBART-Adapt, we plot validation set ROUGE-1 scores against epochs for two previous experiments (high-resource zh$\rightarrow$en and low-resource ja$\rightarrow$en) in Figure~\ref{fig:convergence_plot}. Plotting stops when validation does not improve. We measure convergence in terms of epochs, rather than wall-time. In our experiments, we find that wall-time per epoch for mBART-FT is about merely 1.5 times that for mBART-Adapt, since validation takes a large portion especially when the dataset is small.

\input{convergence_plots.tex}

As Figure~\ref{fig:convergence_plot}(a) shows, with sufficient resources, mBART-FT and mBART-Adapt started with similar ROUGE scores, then the gap quickly increases, suggesting a faster and better convergence rate for fine-tuning. We also observe that mBART-FT converged within fewer epochs. Furthermore, Figure~\ref{fig:convergence_plot}(b) suggests that, in a low-resource condition, even though mBART-FT surpasses mBART-Adapt in terms of ROUGE, they both have similar convergence rates with the gap reduced. These trends indicate that in a high-resource scenario fine-tuning is preferred, whereas in a low-resource scenario, adapters can be used to reduce overhead while maintaining performance.

\section{Domain Adaptation Experiments}
\subsection{Experimental setup}
In addition to multilinguality, we conduct extra experiments on domain adaptation, which is typically tackled using the same pre-training then fine-tuning paradigm. In our setting, we adapt CNN/Daily Mail to XL-Sum, both in English, with various data sizes. Although both datasets are news articles, they differ hugely in writing styles. We start with a BART model \cite{lewis-etal-2020-bart} fine-tuned on the CNN/Daily Mail \cite{nallapati-etal-2016-abstractive} dataset for summarization; it is available as a public model checkpoint.\footnote{\url{https://huggingface.co/ainize/bart-base-cnn}}

To further understand the impact of data availability, we artificially and iteratively make the training data 10 times smaller. This results in five data conditions with sizes ranging from merely 31 to 306.5k. We make sure that larger training splits are supersets of the smaller splits. The validation and test sets remain unchanged at 11.5k as provided in the original dataset. In addition to the XL-Sum dataset, which is in the news domain, we also experimented with adapting CNN/Daily Mail to the BookSum\footnote{\url{https://github.com/salesforce/booksum}} \cite{kryscinski2021booksum} dataset, a collection of narratives from the literature domain such as novels, plays, and stories. Their human written summaries have three levels of granularity, and we use the paragraph-level summaries for our experiment. Unlike the CNN/Daily Mail dataset, we only experiment on the full size of the BookSum dataset.

The English BART checkpoint has in total 139M parameters to be fine-tuned, while adapters have 14.2M parameters (10\%). As an additional parameter-controlled fine-tuning variant, we choose to freeze the entire BART but the last decoder layer, which has 9.5M parameters. The final decoder layer makes up 7\% of the entire model, and has a comparable amount of trainable parameters to an adapter. Similar to the previous setting, we use the Adam optimizer with a learning rate of 1e-5 for BART-FT, and 1e-4 for BART-Adapt. We use a batch size of 4 on XL-Sum, and 8 on BookSum. All other hyperparameter settings are identical to those in the language adaptation experiment.

\begin{table}[tb]
\centering
\setlength{\tabcolsep}{0.97ex}
\begin{tabular}{|c|c|r|c|c|c|c|c|c|c|c|c|}
\hline
 \multirow{2}{*}{Domain} & \multicolumn{2}{c|}{\multirow{2}{*}{Data Size}}
 & \multicolumn{3}{c|}{BART-FT} 
 & \multicolumn{3}{c|}{BART-Adapt} & \multicolumn{3}{c|}{BART-FT-LastLayer} \\
\cline{4-12}
 & \multicolumn{2}{c|}{} & R1 & R2 & RL & R1 & R2 & RL & R1 & R2 & RL \\
\hline
\multirow{5}{*}{XL-Sum} & original & 306.5k & \textbf{34.48}&\textbf{14.73}&\textbf{28.93} & 32.94&13.46&27.60 & 30.20&11.69&25.17 \\
 & medium & 30.65k & \textbf{30.63}&\textbf{11.38}&\textbf{25.31} & 30.15&11.10&25.05 & 26.70&8.67&21.94 \\
 & small & 3065   & 27.27&\textbf{8.91}&\textbf{22.27} & \textbf{27.32}&8.79&22.20 & 23.06&6.21&18.76\\
 & tiny & 307    & 24.10&\textbf{6.52}&19.38 & \textbf{24.29}&6.41&\textbf{19.50} & 19.13&4.12&15.54 \\
 & micro & 31     & 19.69&4.26&15.73 & \textbf{20.74}&\textbf{4.65}&\textbf{16.45} & 16.30&2.20&11.43 \\
\hline
 BookSum & \multicolumn{2}{c|}{111.6k} & \textbf{20.27} & \textbf{4.01} & {15.50} & 20.22  & 3.95 & \textbf{15.57} & 19.33 & 3.56 & 14.93 \\
 \hline
\end{tabular}
\caption{Results for domain adaptation from CNN/Daily Mail to XL-Sum on English (top) with artificially constrained data sizes, and to BookSum (bottom) with full data size.}
\label{tab:domain-xlsum}
\end{table}

\subsection{Results}
We report the experiment results in Table~\ref{tab:domain-xlsum}. The pattern is that for medium to large CNN/Daily Mail data sizes, BART-FT outperforms BART-Adapt significantly. The two methods tie at around 300-3000 training sizes. BART-Adapt wins notably when there are only a handful of examples. This implies that adapters only stand out when the amount of data is extremely limited. In this case, we doubt the importance of training efficiency in adapters when the data size is so small. Instead, we argue that a potential benefit of using adapters is to reduce overfitting. As for BookSum, we can observe that numbers are very similar for both models with BART-FT slightly outperforming BART-Adapt. We argue adapters can do well in domain adaption despite the domain difference as long as there are sufficient training data. Finally, we notice the performance of fine-tuning only the last decoder layer is nowhere near BART-FT or BART-Adapt; this implies the practicability of adapters in summarization.

\begin{table}[t]
% \small
\centering
\scalebox{0.8}{
\begin{tabular}{m{19.2cm}}
\Xhline{1pt}
\textbf{Article:}Lewis Williams, 20, died on 11 January from a shotgun wound suffered in Wath Road, Mexborough. South Yorkshire Police said two men aged 20 and 49 were arrested on Friday in connection with his death, bringing the total number of arrests to eight \ldots 
\\
\textbf{Gold Summary: }Two more people have been arrested in connection with a fatal shooting.
\\
\textbf{BART-FT Summary: }Two more people have been arrested in connection with the fatal shooting of a man in South Yorkshire.
\\
\textbf{BART-Adapt Summary: }{\color{red} \textit{Eight}} more people have been arrested in connection with the death of a man in South Yorkshire.
\\
\hline
\textbf{Article: }BBC News Officials say the country's Olympic Committee will ``oversee participation of women athletes who can qualify''. The decision will end recent speculation as to whether the entire Saudi team could have been disqualified on grounds of gender discrimination \ldots For the desert kingdom, the decision to allow women to compete in the Olympics is a huge step, overturning deep-rooted opposition from those opposed to any public role for women \ldots
\\
\textbf{Gold Summary: }Saudi Arabia is to allow its women athletes to compete in the Olympics for the first time.
\\
\textbf{BART-FT Summary: }Saudi Arabia is to allow women to compete in next year's Olympic and Paralympic Games.
\\
\textbf{BART-Adapt Summary: }Saudi Arabia is to allow women to take part in the {\color{red} \textit{2012 Winter}} Olympics, officials say.
\\
\hline
\textbf{Article: }The vehicle was seen at about 03:45 BST at the fast food giant's branch in Catterick, North Yorkshire. A 19-year-old man was arrested at the site, a short distance from the local golf club, on suspicion of theft and driving while unfit through drink. Police said it was the ``most unusual job'' of the night but officers managed to ``avoid a high-speed pursuit'' \ldots
\\
\textbf{Gold Summary: }A stolen golf buggy was seized after being spotted at a McDonald's drive-thru.
\\
\textbf{BART-FT Summary: }A suspected stolen car was spotted at a McDonald's drive-thru.
\\
\textbf{BART-Adapt Summary: }A man has been arrested after a car was seen driving into a McDonald's branch.
\\
\Xhline{1pt}
\end{tabular}
}
\caption{\label{tab:qualitative_analysis} Examples of gold and generated summaries (from models trained on the full dataset) with their corresponding articles selected from the XL-Sum (English) dataset. Summary phrases italicized and highlighted in red denote hallucinations.}
\end{table}

\begin{table}[ht]
\centering
\begin{tabular}{|c|c|c|c|c|c|}
\hline
Task                   & Data Size & Model      & \multicolumn{1}{c|}{R1} & \multicolumn{1}{c|}{R2} & RL    \\ \hline
\multirow{2}{*}{DialogSum}      & \multirow{2}{*}{12.5k}          & {BART-FT}    & \textbf{47.40}                   & \textbf{24.66}                   & \textbf{39.03} \\ 
                       &  & BART-Adapt & 47.24                   & 24.57                   & 38.56 \\ \hline
\multirow{2}{*}{SAMSum}  & \multirow{2}{*}{14.7k} & BART-FT    & \textbf{49.52}                   & \textbf{24.91}                   & 40.64 \\ 
                      &   & BART-Adapt & 49.38                   & 24.69                   & \textbf{40.99} \\ \hline
\end{tabular}
\caption{Results for task adaption from CNN/Daily Mail to DialogSum and SAMSum.}
\label{tab:task-adaption}
\end{table}

\begin{table}[ht!]
\centering
\begin{tabular}{|c|c|c|c|c|c|}
\hline
Task           & Data Size         & Model      & \multicolumn{1}{c|}{R1} & \multicolumn{1}{c|}{R2} & RL    \\ \hline
\multirow{2}{*}{DialogSum}    & \multirow{2}{*}{12.5k}             & BART-FT*    & 35.60                   & 16.59                   & 29.69 \\ 
                      &  & BART-Adapt* & \textbf{36.35}                  & \textbf{17.03}                   & \textbf{30.25} \\ \hline
\multirow{2}{*}{SAMSum} & \multirow{2}{*}{14.7k} & BART-FT**    & \textbf{40.91}                  & \textbf{14.82}                   & \textbf{32.32} \\ 
                       & & BART-Adapt** & 40.42                  & 14.65                  & 32.28 \\ \hline
\end{tabular}
\caption{Results for robustness analysis of task adaption experiments. Results are directly obtained by using the trained model from the other task without any further training. *denotes the model trained on SAMSum, and **denotes the model trained on DialogSum.}
\label{tab:task-adaption-robust}
\end{table}

\subsection{Qualitative analysis}
To understand the quality of generated summaries between BART-FT and BART-Adapt, we examined a set of randomly selected model outputs from the XL-Sum dataset. We show some examples in Table~\ref{tab:qualitative_analysis}. We find that summaries generated by the two models are roughly the same in terms of informativeness, grammaticality, and fluency. Despite summaries being similar in these aspects, we find that BART-Adapt summaries are more prone to hallucinations, which is a well-known problem in abstractive summarization that summaries are not factual with respect to the source or general knowledge.

\section{Task Transfer Experiments}
\subsection{Experimental setup} In previous settings, we conduct experiments with the fine-tuning paradigm on the subject of language and domain adaption. We also conduct experiments on task adaption to further verify our findings. In particular, we experiment with adapting a news summarization model to dialogue summarization. Dialogue summarization is often considered a much different task from monologic texts (e.g. news in our case) summarization due to its unique challenges. \newcite{chen-etal-2021-dialogsum-challenge} point out that: information flow is reflected in the dialogue discourse structures, summaries are required to be objective, and dialogue is acted at the pragmatic level. For these reasons, we choose to work with the DialogSum \cite{chen-etal-2021-dialogsum-challenge} and the SAMSum \cite{gliwa-etal-2019-samsum} datasets. We follow the previous setting and start with a BART model already fine-tuned on the CNN/Daily Mail dataset, then further train the model on these two datasets separately. We use a batch size of 8 for both DialogSum and SAMSum. All other hyperparameter settings are identical to those in the domain adaptation experiment.

\subsection{Results}
We report the experiment results in Table~\ref{tab:task-adaption}. We can observe that despite the dataset, BART-FT almost always beats BART-Adapt. However, we can notice that the performance gap is rather small, possibly due to the small dataset sizes. This is consistent with our earlier findings that adapters are on par with fine-tuning when the amount of training data is limited.

\subsection{Model robustness} In addition to model performance, we also examine the robustness of models with either fine-tuning or adapters. In particular, we evaluate the model in a zero-shot manner where we directly test the DialogSum model on the SAMSum dataset, and vice versa. We present the results in Table~\ref{tab:task-adaption-robust}. We can first observe that performance drops significantly compared to those in Table~\ref{tab:task-adaption}. Moreover, BART-Adapt has better performance than BART-FT on the DialogSum dataset, and it achieves very similar results on the SAMSum dataset. This suggests that adapters are more robust in a zero-shot setup with fewer data; the reason could be less overfitting introduced by a limited number of parameters in adapters.

\input{datasize_vs_rouge_percent.tex}

\subsection{Effect of data availability on performance}

Our results suggest that fine-tuning generally surpasses adapters under all three settings (language, domain, and task adaption). In addition, we observe that the amount of training data affects the performance gap between the two methods. To further validate this observation, we plot the percentage change in ROUGE performance (between those of fine-tuning and those of adapters) against the training size (log-scale) and we provide the visualization in Figure~\ref{fig:datasize_plot}. We use the average number of ROUGE-1/2/L to represent the performance. From the plot, we can see that percentage change in ROUGE has an obvious positive relationship with the training data size which means that as the amount of training data increases, the performance gap between BART-FT and BART-Adapt increases as well. Looking at the tasks individually, we can see that for language adaption tasks with relatively small amounts of data, this trend is not very notable. The trend is most salient on domain adaption tasks since we manually controlled the data size for the experiment for adapting CNN/Daily Mail to XL-Sum.

\FloatBarrier % all figures, table, etc, must be rendered before the conclusion.
\section{Conclusions and Future Work}
With large PLMs coming to light, we investigate fine-tuning and adapter strategies for transfer learning in abstractive summarization. We demonstrated that the performance gap between the two strategies is positively correlated with the availability of training resources, despite the languages being tested. Further analysis on domain adaptation and task adaption produces agreeing observations. We conclude that for realistically large summarization datasets, full fine-tuning will guarantee the best output quality. On the other hand, when resources are scarce, the advantages of adapters emerge in the niche market.

Most summarization datasets are web-crawled or machine-translated, resulting in non-optimal data quality. We plan to perform more qualitative analysis on the model outputs such as linguistic interpretation and human evaluation. In addition, we only experimented with fine-tuning and using adapters on mBART and BART for abstractive summarization, so there is room for research on other large PLMs, as well as other NLP tasks in the future.

\section*{Acknowledgements}
We thank the reviewers of the paper for their feedback. Zheng Zhao is supported by the UKRI Centre for Doctoral Training in Natural Language Processing (grant EP/S022481/1). Pinzhen Chen is supported by a donation to Kenneth Heafield. This work does not necessarily reflect the opinion of the funders.

\bibliographystyle{ccl}
\bibliography{custom}

\appendix

\newpage
\section{Model Configurations}
\label{sec:appendix-hyper-parameters}

We tuned the hyperparameters using the validation set. We list the hyperparameters in Table~\ref{tab:config}, and highlight the selected ones in bold if multiple values are tried out. Instead of an expensive grid search on all combinations, we searched for the best configurations one by one. We performed a single run for each experiment.

\begin{table}[htb]
\centering
\begin{tabular}{|l|l|} 
\hline
\textbf{Configuration}  & \textbf{Value} \\
\hline
 training toolkit       & PyTorch \cite{NEURIPS2019_9015} \\
 stopping criterion     & validation ROUGE \\
 learning rate          & 1e-3, 5e-3, \textbf{1e-4} (mBART+Adapt), 5e-4, \textbf{1e-5} (mBART-FT), 5e-5 \\
 optimizer              & Adam \cite{KingmaB14} \\
 beta1, beta2           & 0.9, 0.999 \\
 weight decay           & 1e-6 \\
 loss function          & cross-entropy \\
 decoding batch size    & 1 \\
 decoding beam size     & 5 \\
 decoding len. penalty   & 1.0 \\
\hline
 adapter reduction factor      & 1, \textbf{2}, 8, 16 \\
\hline
 \textit{trainable} parameters   & mBART-FT: 610M \\ 
                        & mBART-Adapt: 50M \\
\hline
\end{tabular}
\caption{Model and training configurations.}
\label{tab:config}
\end{table}

\end{document}

%% file: convergence_plots.tex
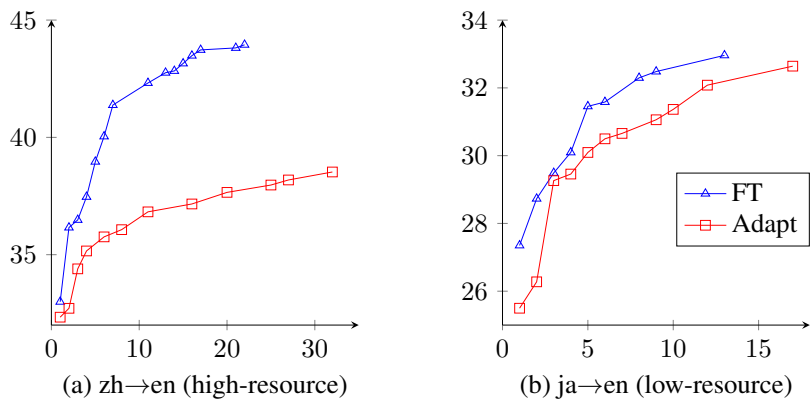
\begin{figure}[ht!]
\centering
\resizebox{0.3\textwidth}{!}{%
\begin{tikzpicture}
\begin{groupplot}[
    group style={
        group name=zh-en,
        group size=1 by 2
    },
    width=6cm,
    xmin=0, xmax=35,
    xlabel=(a) zh$\rightarrow$en (high-resource)
]

\nextgroupplot[ymin=32,ymax=45,
               axis y line=left,
               axis x line=bottom,
               height=6cm]

\addplot[
    color=blue,
    mark=triangle,
    ]
    coordinates {
    (1,32.9927)(2,36.1564)(3,36.4823)(4,37.4642)(5,38.9624)(6,40.0398)(7,41.3765)(11,42.3185)(13,42.7553)(14,42.8314)(15,43.1508)(16,43.4828)(17,43.7284)(21,43.8133)(22,43.9464)
    };
    % \addlegendentry{3-task};
\addplot[
    color=red,
    mark=square,
    ]
    coordinates {
    (1,32.3361)(2,32.7142)(3,34.3963)(4,35.1592)(6,35.7592)(8,36.0669)(11,36.826)(16,37.1616)(20,37.6556)(25,37.9686)(27,38.1847)(32,38.525)
    };
    % \addlegendentry{5-task};

\end{groupplot}
\end{tikzpicture}
}%
\hspace{6ex}
\resizebox{0.3\textwidth}{!}{%
\begin{tikzpicture}
\begin{groupplot}[
    group style={
        group name=ja-en,
        group size=1 by 2
    },
    width=6cm,
    xmin=0, xmax=18,
    xlabel=(b) ja$\rightarrow$en (low-resource),
    legend cell align={left},
    legend style={ at={(1,0.5)}, anchor=north east}
]

\nextgroupplot[ymin=25,ymax=34,
               axis y line=left,
               axis x line=bottom,
               height=6cm]

\addplot[
    color=blue,
    mark=triangle,
    ]
    coordinates {
    (1,27.35)(2,28.73)(3,29.49)(4,30.096)(5,31.4557)(6,31.5834)(8,32.2961)(9,32.4819)(13,32.9601)
    };
    \addlegendentry{FT};
\addplot[
    color=red,
    mark=square,
    ]
    coordinates {

    (1,25.4972)(2,26.2765)(3,29.2632)(4,29.4568)(5,30.0944)(6,30.4972)(7,30.6562)(9,31.0596)(10,31.3657)(12,32.0804)(17,32.6414)
    };
    \addlegendentry{Adapt};

\end{groupplot}

\end{tikzpicture}
}%
\caption{Validation ROUGE-1 (y-axis) against epochs (x-axis) for mBART-FT and mBART-Adapt in different data conditions.}
\label{fig:convergence_plot}
\end{figure}

%% file: datasize_vs_rouge_percent.tex
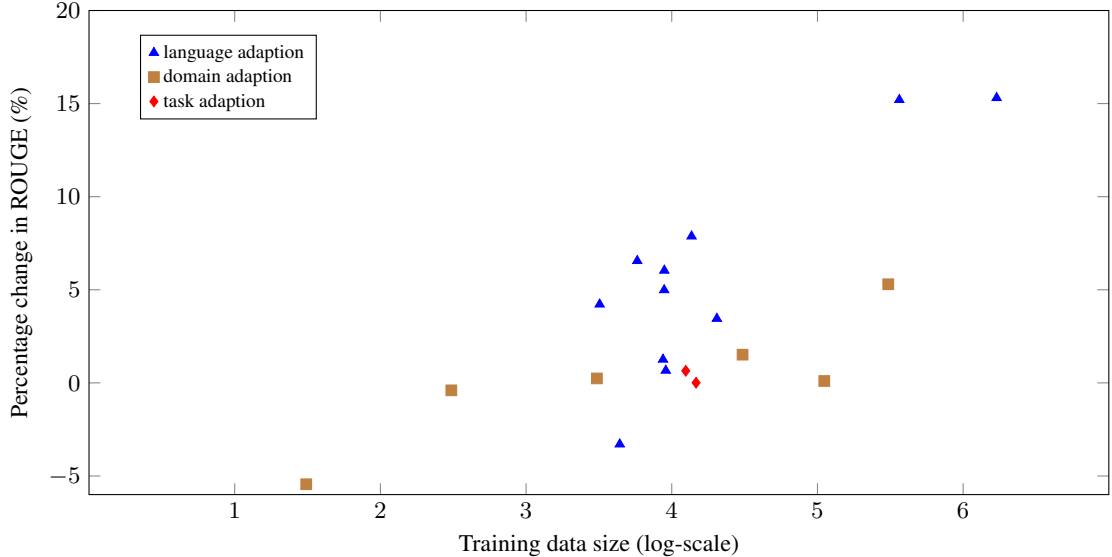
\begin{figure*}[ht!]
\small
\centering
    \begin{tikzpicture}
    \begin{axis}[
        ytick={-5,0,5,10,15,20},
        xtick={1,2,3,4,5,6},
        xmax=7,
        xmin=0,
        ymax=20,
        ymin=-6,
        ylabel=Percentage change in ROUGE (\%),
        xlabel=Training data size (log-scale),
        height=8cm,
        width=15cm,
        /pgf/number format/fixed,
        /pgf/number format/precision=2,
        domain=0:7,
        legend cell align={left},
        legend style={ at={(0.05,0.95)}, anchor=north west, nodes={scale=0.8, transform shape}}
        ]
    \addplot[mark=triangle*,blue,only marks,]
    table {
        x y  
        6.23 15.307
        5.562 15.202
        4.3096 3.45
        4.136 7.873
        3.948 4.988
        3.949 6.036
        3.959 0.663
        3.94 1.257
        3.763 6.554
        3.643 -3.297
        3.505 4.218
    };
    \addlegendentry{language adaption};
    \addplot[mark=square*,brown,only marks,]
    table {
        x y  
        5.486 5.298
        4.486 1.515
        3.486 0.240
        2.487 -0.4
        1.491 -5.444
        5.047 0.101
    };
    \addlegendentry{domain adaption};
    \addplot[mark=diamond*,red,only marks,]
    table {
        x y  
        4.096 0.648
        4.167 0.009
        
    };
    \addlegendentry{task adaption};
    \end{axis}
    \end{tikzpicture}
    \caption{The effect of the training data size on ROUGE difference between the fine-tuning and adapter strategy. We display how much percent FT is better than using adapters. Note that data points from different tasks (with different shapes and colors) are not strictly comparable.}\vspace{-1ex}
    \label{fig:datasize_plot}
\end{figure*}